\documentclass[runningheads]{llncs}

\usepackage{graphicx}
\usepackage{multirow}
\usepackage{amsmath}
\begin{document}
\title{A novel tactile palm for robotic object manipulation}
%
%
\author{Fuqiang Zhao\inst{1} \and
Bidan Huang\inst{2} \and
Mingchang Li\inst{3} \and
Mengde Li\inst{4} \and
Zhongtao Fu\inst{5} \and
Ziwei Lei\inst{1} \and
Miao Li\inst{1,4}
}
\authorrunning{F. Zhao et al.}
%
\institute{The School of Power and Mechanical Engineering, Wuhan University, China \and
Tencent Robotics X, Shenzhen, Guangdong Province, China \and
Department of Neurosurgery, Renmin Hospital of Wuhan University, China\\\and
The Institute of Technological Sciences, Wuhan University, China\\ \and
School of Mechanical and Electrical Engineering, Wuhan Institute of Technology, Wuhan, China \\
\email{miao.li@whu.edu.cn}
}
%
\maketitle              
\begin{abstract}
Tactile sensing is of great importance during human hand usage such as object exploration, grasping and manipulation. 
Different types of tactile sensors have been designed during the past decades, which are mainly focused on either the fingertips for grasping or the upper-body for human-robot interaction. 

In this paper, a novel soft tactile sensor has been
designed to mimic the functionality of human palm that can estimate the contact state of different objects. The tactile palm mainly consists of three parts including an electrode array, a soft cover skin and conductive sponge. The design principle are described in details, with a number of experiments showcasing the effectiveness of the proposed design.
\footnote[1]{This work was supported by Suzhou Key Industry Technology Innovation Project under the grant agreement number SYG202121 and by the Fundamental Research Funds for the Central Universities under the grant agreement number 2042023kf0110. (Corresponding author:  Miao Li.)}

\keywords{Tactile Sensor \and Robotic Grasping and Manipulation \and Robot Tactile Palm.}
\end{abstract}

\section{Introduction}
Tactile sensing is one of the most important modalities that endows the human hands with incomparable dexterity to explore, grasp and manipulate various objects \cite{jones2006human,castiello2005neuroscience,billard2019trends}. These tasks involve physical contacts with the real world, which requires the tactile information to guide the sequence of actions for the task accomplishment \cite{johansson2009coding}. Taking robotic grasping as an example, vision can inform the robot about the placement of the hand and the finger configuration, but ultimately the tactile information is still required to predict the contact state in terms of contact locations, contact forces and the grasp stability as well.

During the past decades, a large variety of tactile sensors have been designed for different robotic applications \cite{bartolozzi2016robots,dahiya2009tactile,kappassov2015tactile}, which can be roughly categorized into two groups: the tactile sensors installed on the fingertips for robotic grasping and dexterous manipulation \cite{yousef2011tactile,luo2017robotic}, and the tactile (artificial) skin for human-robot interaction \cite{argall2010survey,silvera2015artificial}. The former group requires the tactile sensor to be compact and with high-resolution, similar to the human fingertips \cite{johansson2009coding}. Conversely, the latter group is usually used for human-robot interaction, where only a limited number of interaction patterns are required to be recognized in a human-like manner. Therefore the tactile skin must be soft and even stretchable in order to cover a large area with the compromise of the spatial resolution.   

\begin{figure}[!h]
    \centering
    \includegraphics[width=0.5\linewidth]{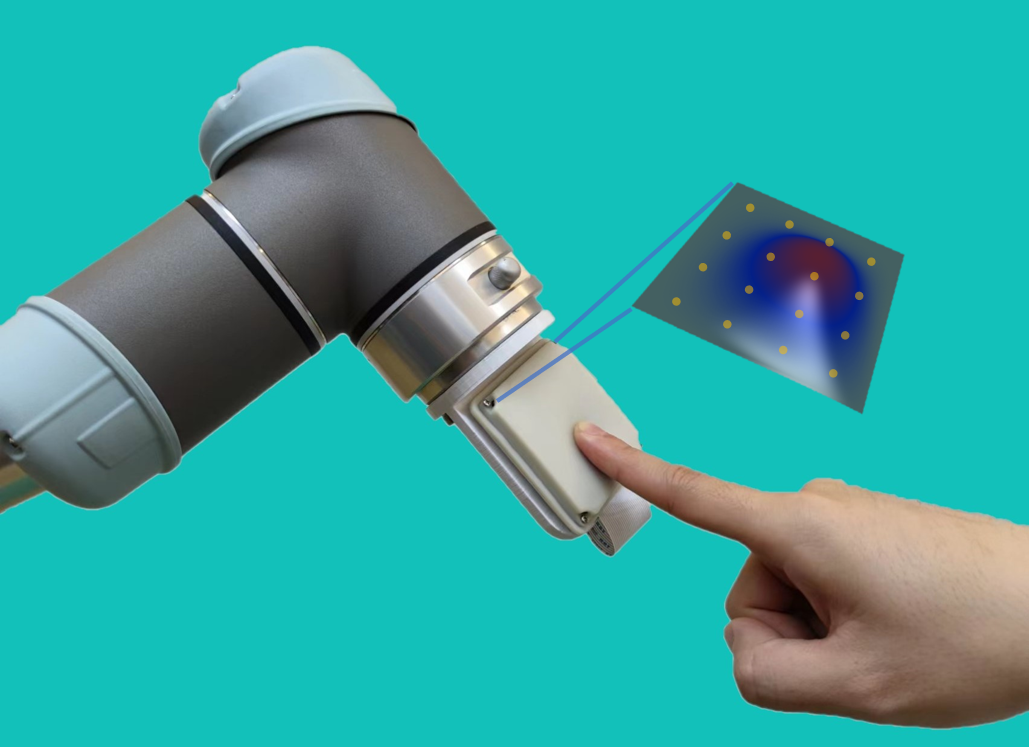}
    \caption{The robotic tactile palm mounted on the UR5 robot arm. The top right window shows the tactile response to the corresponding contact state, where the filled yellow circles represent the positions of the 16 sensing electrodes.}.
    \label{System}
    \vspace{-1cm}
\end{figure}

To summarize, previous studies of tactile sensors in robotics have focused mostly on the fingertips sensing and the tactile skin for the humanoid robot body. However, the palm is also an indispensable part for object manipulation, particularly in-hand manipulation \cite{castiello2005neuroscience,stival2019quantitative}. In this work, we propose a biomimetic design of an inexpensive robotic tactile palm that can provide the object contact information such as contact position and contact forces during in-hand manipulation. The tactile palm comprises three main components: an array of sensing electrodes, a soft cover skin and a conductive sponge, as shown in Fig.~\ref{System}.    

\subsection{Robot Palm for Soft Grasping}


The superior performance of soft hands is achieved at the expense of dexterity and precision. Many recent works have developed active palms to improve the dexterity of soft hand through implementing an additional degree of freedom to the palm \cite{wang2021novel,capsi2020exploring,yamaguchi2012robot,meng2020tendon,subramaniam2020design,sun2020design,pagoli2021soft}. For example, the palm-object interaction is actively controlled through varying the friction force by either changing the coefficient of friction or the normal load \cite{teeple2022controlling}. Jamming-based palm has been designed to actively adapt its shape for different objects geometry with variable stiffness \cite{lee2021soft,li2019novel}. While these new designs can help the soft hands to grasp more diverse objects with better stability, they still lack of the ability to estimate the contact state between the object and the hand. In another word, the grasp stability is passively guaranteed by the mechanical design of the hand.    
 
\subsection{Tactile Palm for In-hand Object Manipulation}
For in-hand object manipulation, the goal is to move the object to a desired state rather than to secure the object.
To this end, it is extremely important to acquire the information about the object contact state. Previous studies have mostly focused on using vision and hand proprioception to estimate the object state or hand-object configuration \cite{choi2016using,andrychowicz2020learning,hang2020hand}. This is partly due to the difficulty to design and integrate inexpensive but robust tactile sensor with the hand. 
Tactile dexterity is proposed for dexterous manipulations that render interpretable tactile information for control \cite{hogan2020tactile,hogan2018tactile}. The tactile pad used for manipulation is largely adapted from two vision-based tactile sensors that use a camera to measure tactile imprints-GelSlim \cite{donlon2018gelslim} and GelSight \cite{yuan2017gelsight}. The vision-based tactile sensors have the advantages of low-cost and robust for physical interactions. However, vision-based tactile sensing has two intrinsic issues of non-uniform illumination and a strong perspective distortion, both of which require a careful calibration process.

In this paper, inspired by the work \cite{wettels2008biomimetic} and our previous work\cite{lei2022biomimetic}, we propose a novel design of an inexpensive robotic tactile palm, which can be used to estimate the contact state of the object during grasping and in-hand manipulation.
For the sake of clarity, in this work we present:
\begin{itemize}
\item a novel design and fabrication method of a soft tactile palm consisting of an array of sensing electrodes, a soft cover skin inside and the conductive sponge.
\item a systematic approach to estimate the contact positions and contact forces for the proposed tactile palm.
\end{itemize}


The rest of this paper is organized as follows: The  design details and fabrication process are described in Section 2, while the methods used for contact position and contact forces estimation are given in Section 3. The experimental results are presented in Section 4 to demonstrate the effectiveness of this tactile palm, with a discussion and a conclusion in Section 5. 

\begin{figure}[!h]
    \centering
    \includegraphics[width=0.6\linewidth]{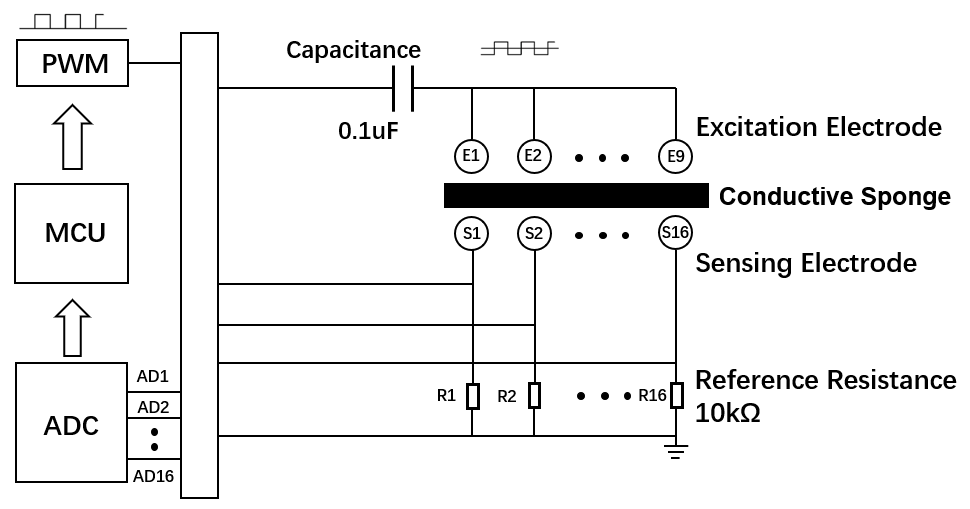}
    \caption{The architecture of the electric circuitry to measure the electrode array impedance.}.
    \label{Principle}
    \vspace{-1cm}
\end{figure} 

\section{Design and Fabrication}
\subsection{Design Principle and Overall Structure}
Similar to the design principle in \cite{wettels2008biomimetic}, the proposed tactile palm consists of a rigid electrode board covered by conductive dielectric contained within an elastomeric skin. Multiple electrodes including the excitation electrodes and the sensing electrodes are mounted on board and connected to an impedance-measuring circuitry as shown in Fig.~\ref{Principle}. 

When external force is applied to the elastomeric skin, the conductive dielectric around the corresponding electrodes will deform. Therefore the measured impedance changes accordingly, which contains information about the external force and contact location as well.

For the conductive dielectric, we selected a conductive sponge composed of polymer, which is a porous composite material that exhibits excellent resilience and conductivity. The conductive sponge is placed around the electrodes. When pressure is applied to the skin, the conductive sponge deforms and alters the conductivity of the adjacent electrode points. The elastomeric skin is made by 3D printing using TPU material, which encapsulates all the devices inside the skin.

\begin{figure}[!h]
    \centering
    \includegraphics[width=0.5\linewidth]{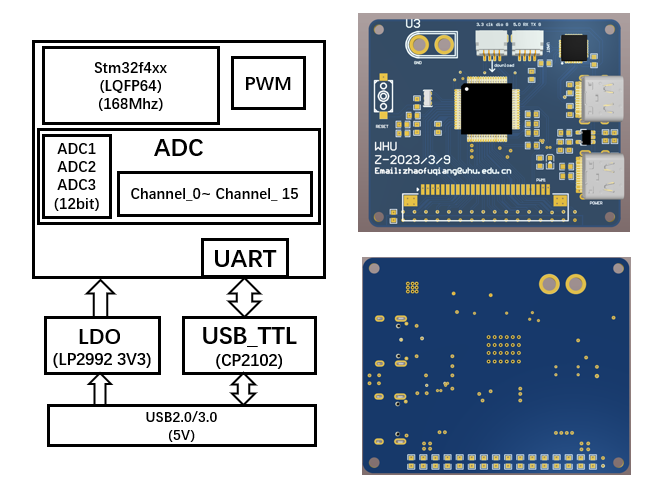}
    \caption{The microcontroller(Stm32F4x) embedded with three 12-bit ADCs. Here, the excitation pulse is given through pulse width modulation (PWM) at a rate of 10000 Hz.}.
    \label{MCU}
    \vspace{-1cm}
\end{figure} 

\subsection{Fabrication and Assembly}
\textbf{Sensor Base and Skin:} Both the base of sensor and the the skin are fabricated using 3D printing with an accuracy of $0.02mm$. (3D printer: UltiMaker Extended+, material: TPU). 

\textbf{Data Acquisition Module:} For the data acquisition and processing module, we chose STM32F4x series chip\footnote{https://atta.szlcsc.com/upload/public/pdf/source/20140801/1457707197138.pdf} to design a simple analog-to-digital conversion (ADC) microcontroller with a sampling rate of 200 Hz. STM32F4 has frequency 168Mhz, embedded with three 12-bit ADCs. Each ADC module shares 16 external channels. Through the interface and soft wire, the microcontroller sends pulse-width modulation (PWM) signals at a frequency of $10000$ Hz after capacitive filtering as an excitation signal to the excitation electrode. The signals collected by the sensing electrodes are then transmitted back to the microcontroller.
The microcontroller uses a USB to TTL level serial module to communicate with the host computer and send the collected voltage values.
The PCB consists of two layers, with most of the devices placed on the top layer. The overall size of the sensor contact acquisition module is 50mm*40mm, as shown in Fig.~\ref{MCU}.

\begin{figure}[!h]
    \centering
    \includegraphics[width=0.4\linewidth]{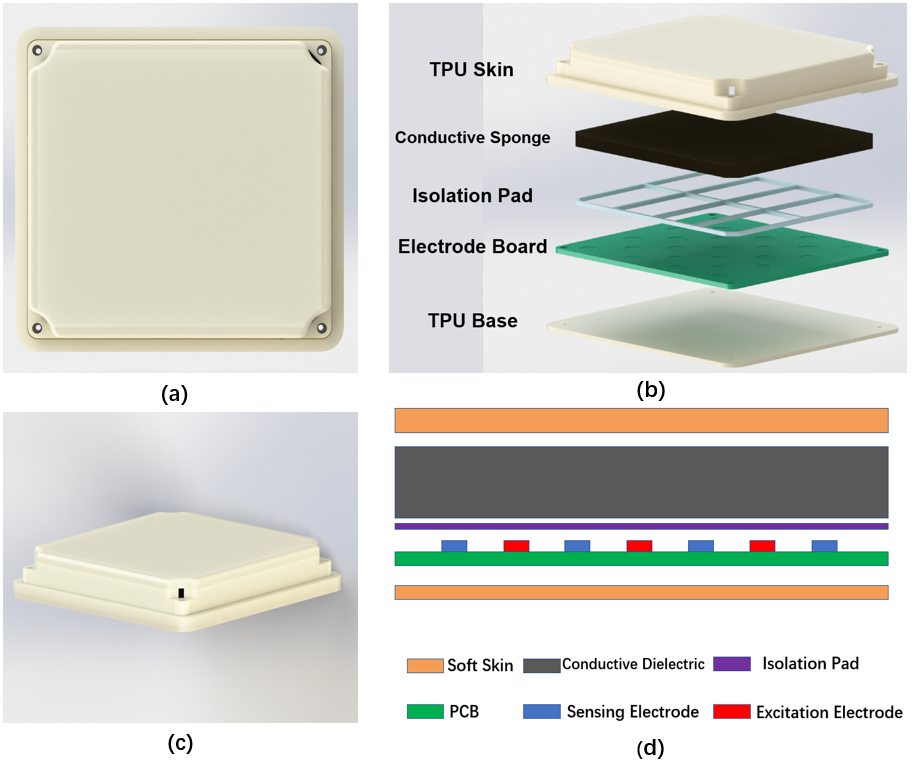}
    \caption{The overall structure of the proposed tactile palm. It consists of an electrode board covered by conductive sponge contained within an elastomeric skin. Multiple electrodes including the excitation electrodes and the sensing electrodes are mounted on board and connected to an impedance-measuring circuitry.}
    \label{assembly}
\end{figure}

\begin{figure}[!h]
    \centering
    \includegraphics[width=0.4\linewidth]{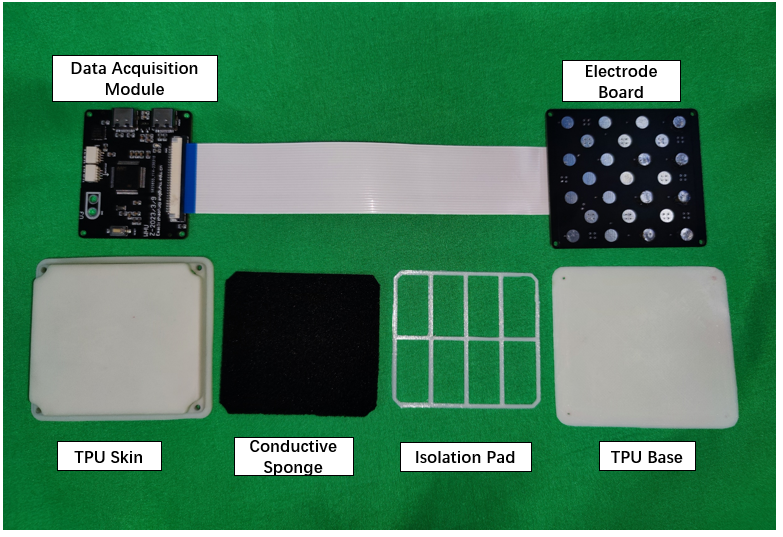}
    \caption{The assembly of the tactile palm.}
    \label{AssemblyP}
\end{figure}

\textbf{Assembly:} The assembly consists of a base, a PCB, a conductive sponge, an isolation pad, and a soft skin, all arranged in layers as shown in Fig.~\ref{assembly} and Fig.~\ref{AssemblyP}. The isolation pad serves the purpose of isolating the conductive sponge from the electrode points on the PCB when no contact force is applied. The total fabrication cost of our tactile palm, including the data acquisition card, is less than 200 dollars.

\begin{figure}[!h]
    \centering
    \includegraphics[width=0.5\linewidth]{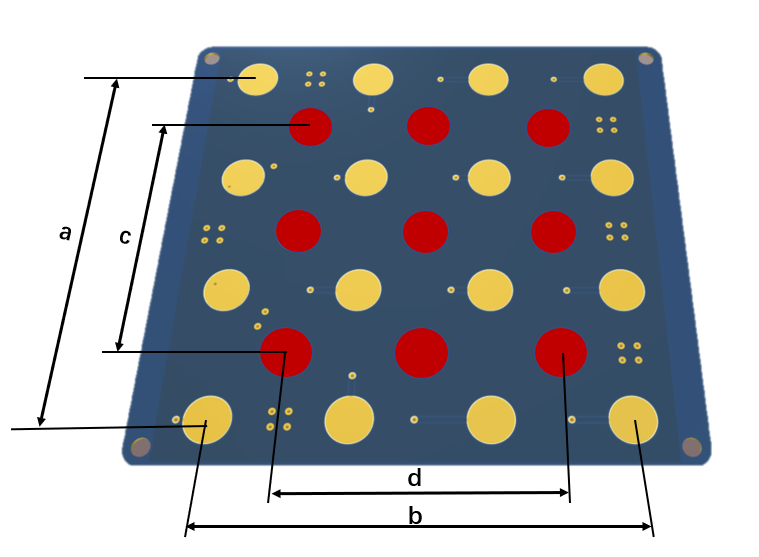}
    \caption{The spatial layout of the sensing electrodes and the excitation electrodes. The design parameters are: a=45mm, b=45mm, c=30mm, d=30mm. Red represent excitation electrode, and yellow represent sensing electrode.}.
    \label{PCB}
    \vspace{-1cm}
\end{figure} 

\section{Contact State Estimation}
\subsection{Contact Position Estimation}
\begin{figure}[!h]
    \centering
    \includegraphics[width=0.6\linewidth]{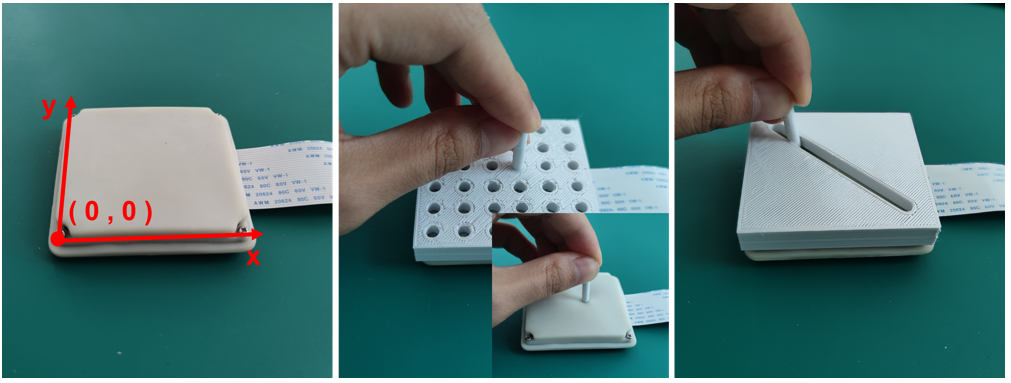}
    \caption{Left: the coordinate system of the tactile palm; Middle: printed model to test the position accuracy at several given points; Right: printed model to test the position accuracy along a given straight line.}.
    \label{sense_test}
    \vspace{0cm}
\end{figure}

\begin{figure}[!h]
    \centering
    \includegraphics[width=0.7\linewidth]{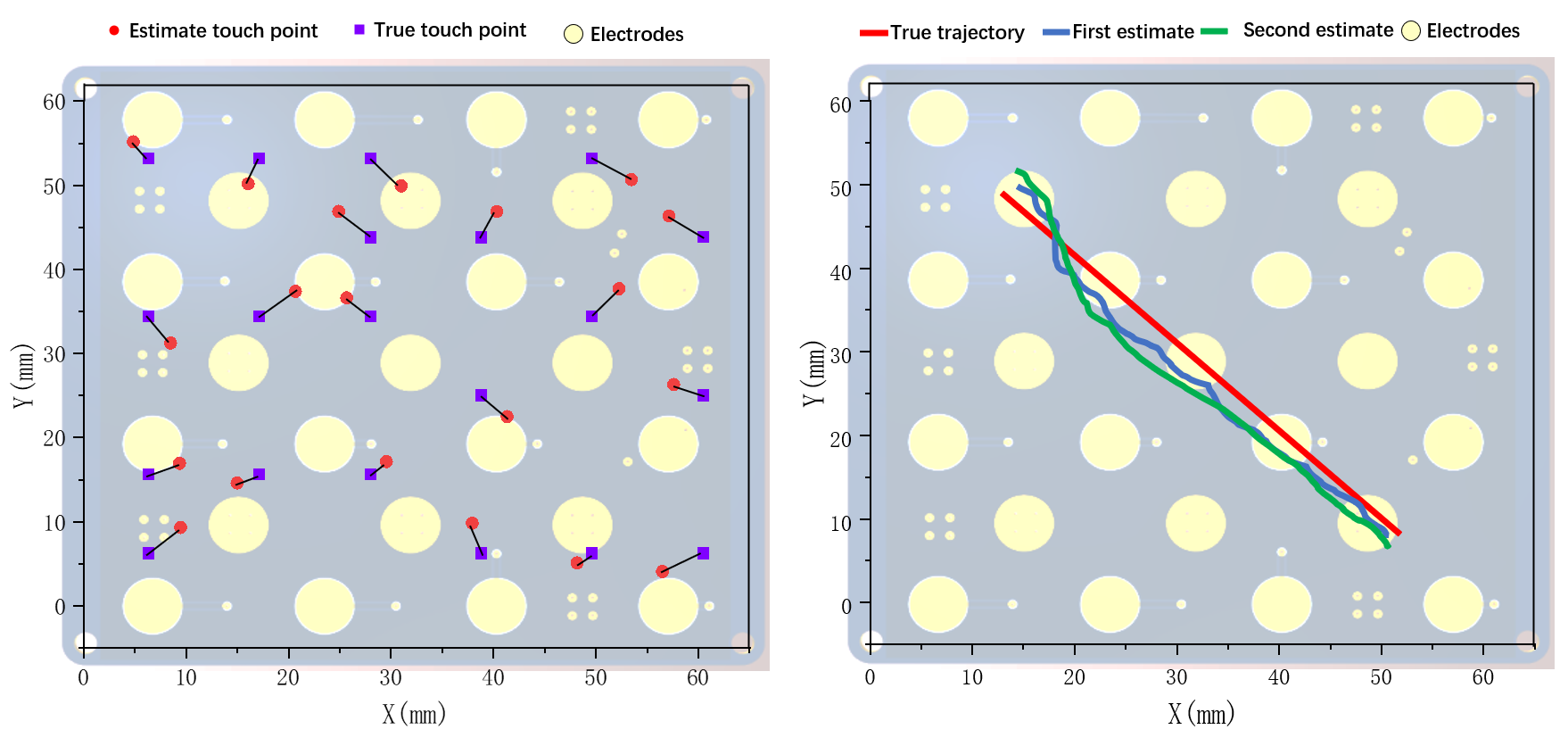}
    \caption{Left: The contact position estimation for several given points. The average accuracy of the estimated position is around 2.7mm and the maximal estimation error of position is 4.5mm; Right: The straight line estimation during three trials. A printed probe is moving along a given slot for two times as shown in Fig.~\ref{sense_test} (right) and the estimated traces are shown with different colors.}.
    \label{sense_result}
    \vspace{-1cm}
\end{figure}

Our tactile palm has $16$ sensing electrodes as shown in Fig.~\ref{PCB}, $\{\mathbf{P}_i\}_{i=1...16}$, which are used as the anchor points to estimate the contact location. Once contacted with objects, $16$ dimensional raw pressure data $S=\{S_i\}_{i=1...16}$ from the $16$ sensing electrodes were sampled at $200$Hz. The contact position is estimated as:
\begin{equation}
\mathbf{P}_c=\sum\limits_{i=1}^{m}\alpha_i\mathbf{P}_i
\end{equation}
where $m$ is the number of activated electrodes that generally locate close to the contact points. $\alpha_i$ is the coefficient of the $i$th electrode, which can be computed as follows:
\begin{equation}
\alpha_i=\frac{\beta_i}{\sum\limits_{1}^{m}\beta_i},~~~~
\beta_i =1-e^{-\frac{2(S_i-S_{bi})^2}{\sigma_i^2}}
\end{equation}
where $S_{bi}$ and $\sigma_i$ are the baseline of pressure and range of variation for the $i$th electrode, respectively. To test the accuracy of the contact position estimation, we use two 3D printed models as shown in Fig.~\ref{sense_test} to test the position accuracy at several given points as well as along a given straight line. The average accuracy of the estimated contact position is around $2.7$mm in practice and the maximal estimation error of position is $4.5$mm. The position estimation and the straight line estimation is shown in Fig.~\ref{sense_result}.

\subsection{Contact Force Estimation}
\begin{figure}[!h]
    \centering
    \includegraphics[width=0.4\linewidth]{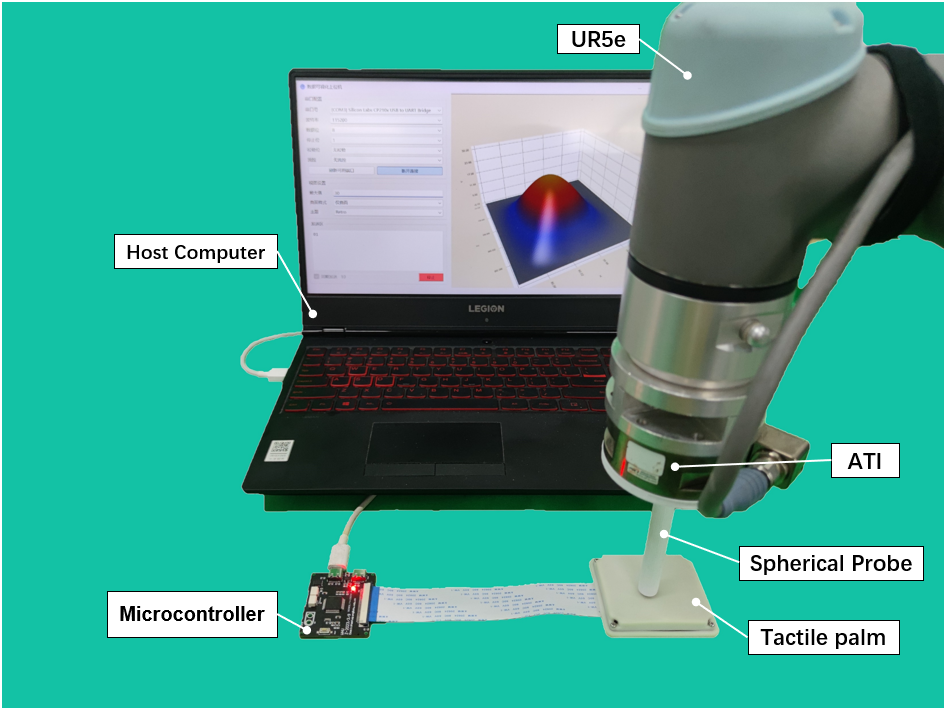}
    \caption{The robot setup for the contact force estimation. The robot is mounted with a force/torque sensor and equipped with a spherical probe (diameter: 5mm)}.
    \label{force_test}
    \vspace{-1cm}
\end{figure}

The tactile palm can provide $16$ dimensional raw pressure data $\{S_i\}_{i=1...16}$ from $16$ sensing electrodes distributed as in Fig.~\ref{PCB}. The goal of contact forces estimation is to predict the contact forces from these raw data. To this end, an ATI net Force/Torque sensor is used to measure the contact forces $\{F_i\}_{i=1...3}$ between the tactile palm and the force sensor, while the contact forces are varied by the robot to take all the possible contact situations into account. The setup for the contact force estimation is shown in Fig.~\ref{force_test}.

We uniformly sampled $28$ points on the tactile palm. For each point, we first move the robot to a home position that is $20$mm above the sampled point. Then we move the robot downwards until the contact is detected by the tactile sensor. Then the robot is moving downwards for a distance of $4$mm in position control mode and both the force and tactile data are collected during this pressing procedure. Finally, the robot moves upwards to the home position. This process is repeated $5$ times for each points.

The sampling rates for both tactile palm and force sensor are down-sampled to $100$ Hz. $28\times5\times700=98000$ data points are collected as the training dataset. In addition, we randomly sampled another $10$ points and used the same procedure to collect $7000$ data points as the testing dataset.  
For simplicity, the training data set is denoted by $\{X_i^j=[S^j,F^j]\}_{i=1...27}^{j=1...98000}$. In principle , any nonlinear regression method can be used to learn the mapping from $S$ to $F$. The Gaussian Mixture Model (GMM) is adopted here due to its flexibility in modeling density function for high dimensional data.

The joint probability density function of $S$ and $F$ is estimated by a GMM composed of $K$ Gaussian functions:   
\begin{equation}
p(X)=\sum\limits_{k=1}^{K}\pi_k\mathcal{N}(X|\boldsymbol\mu_k, \boldsymbol\Sigma_k)
\label{equation::GMM}
\end{equation}
where $\pi_k$ is the prior of the $k$th Gaussian component and $\mathcal{N}(\boldsymbol\mu_k, \boldsymbol\Sigma_k)$ is the Gaussian distribution with mean $\boldsymbol\mu_k$ and covariance $\boldsymbol\Sigma_k$. The number of Gaussian functions $K$ is selected by Bayesian information criterion (BIC), in this work $K=7$. Other parameters $\{\pi_k, \boldsymbol\mu_k, \boldsymbol\Sigma_k\}$ are learned using EM algorithm.

\begin{figure}[!h]
    \centering
    \includegraphics[width=0.5\linewidth]{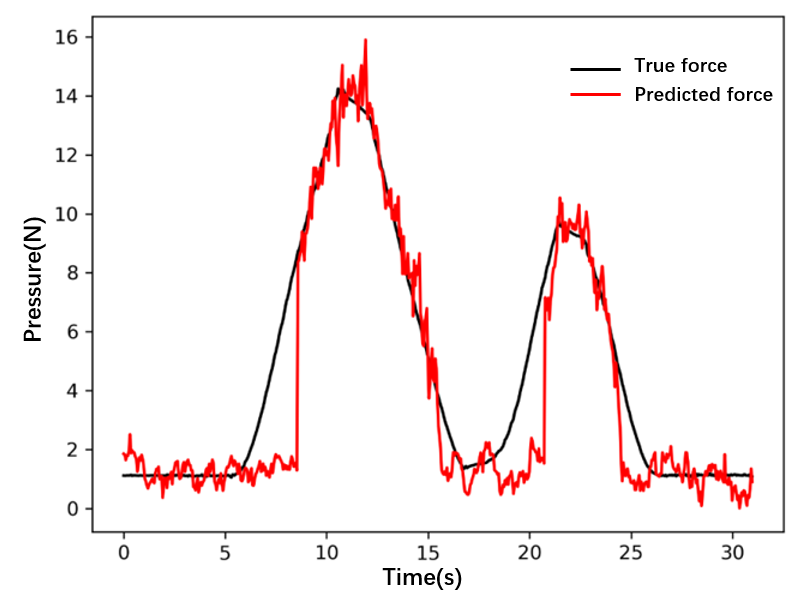}
    \caption{The force calibration along the normal direction. The Root Mean Square Error (RMSE) of the estimated contact force is 0.38N. Force Unit: Newton.}.
    \label{force_result}
    \vspace{-1cm}
\end{figure}

With the trained model, now given a new tactile response $S^*$, the contact forces can be predicted as $F^*=E[F|S^*]$, which can be easily computed using Gaussian Mixture Regression (GMR). The Root Mean Square Error (RMSE) is $0.38$N and the testing result is shown in Fig.~\ref{force_result}.

As a comparison, we also applied SVR (Support Vector Regression) and DNN (Deep Neural Network) for the contact force estimation. The best performance of the estimation in terms of RMSE is $0.54$N for SVR and $0.42$N for DNN, respectively. This can be explained by the fact that SVR and DNN attempt to learn the regression between the force and the tactile data directly, which is a quite strict constraint and is also sensitive to sensory noise. As a comparison, GMM attempts to model the joint density between the contact force and the tactile information, from which the mapping to predict the contact force can be implicitly inferred.

\section{Experiment and Result}
\begin{figure}[!h]
    \centering
    \includegraphics[width=0.6\linewidth]{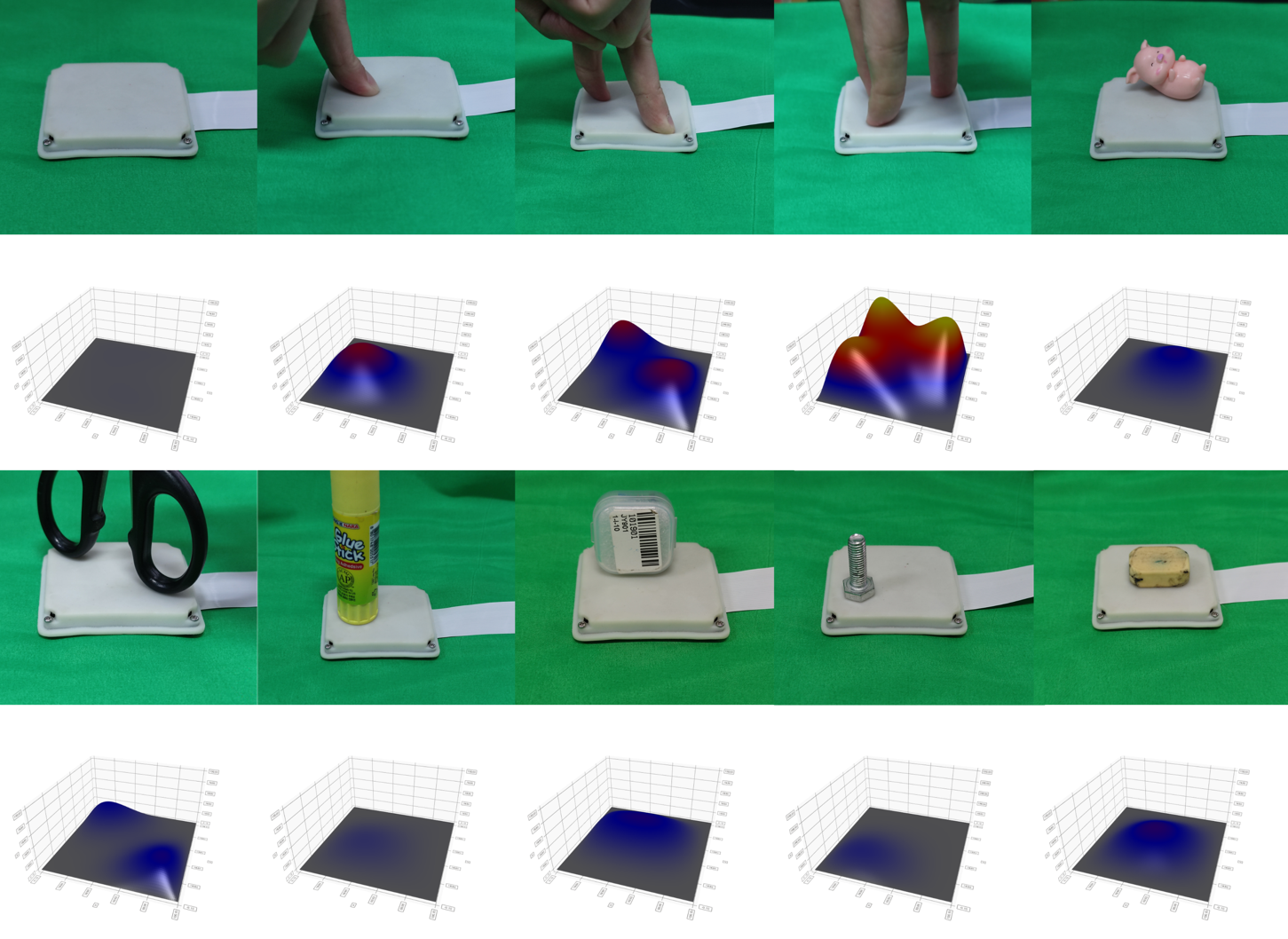}
    \caption{The tactile responses to different objects. From the left to right: touch by one finger, touch by two fingers, touch by three fingers, earphone, stone, clip, charger, screw, solid glue, 3D printed workpiece.}.
    \label{test}
\end{figure}

In the experiment, we tested the tactile response to different everyday objects, as shown in Fig.~\ref{test}. For small and light object like earphone, the proposed tactile palm can still generate a very clear tactile pattern. It is worth mentioning that the sensitivity of the tactile palm can still be improved by using a softer elastomeric skin and more electrodes.

\begin{figure}[!h]
    \centering
    \includegraphics[width=0.6\linewidth]{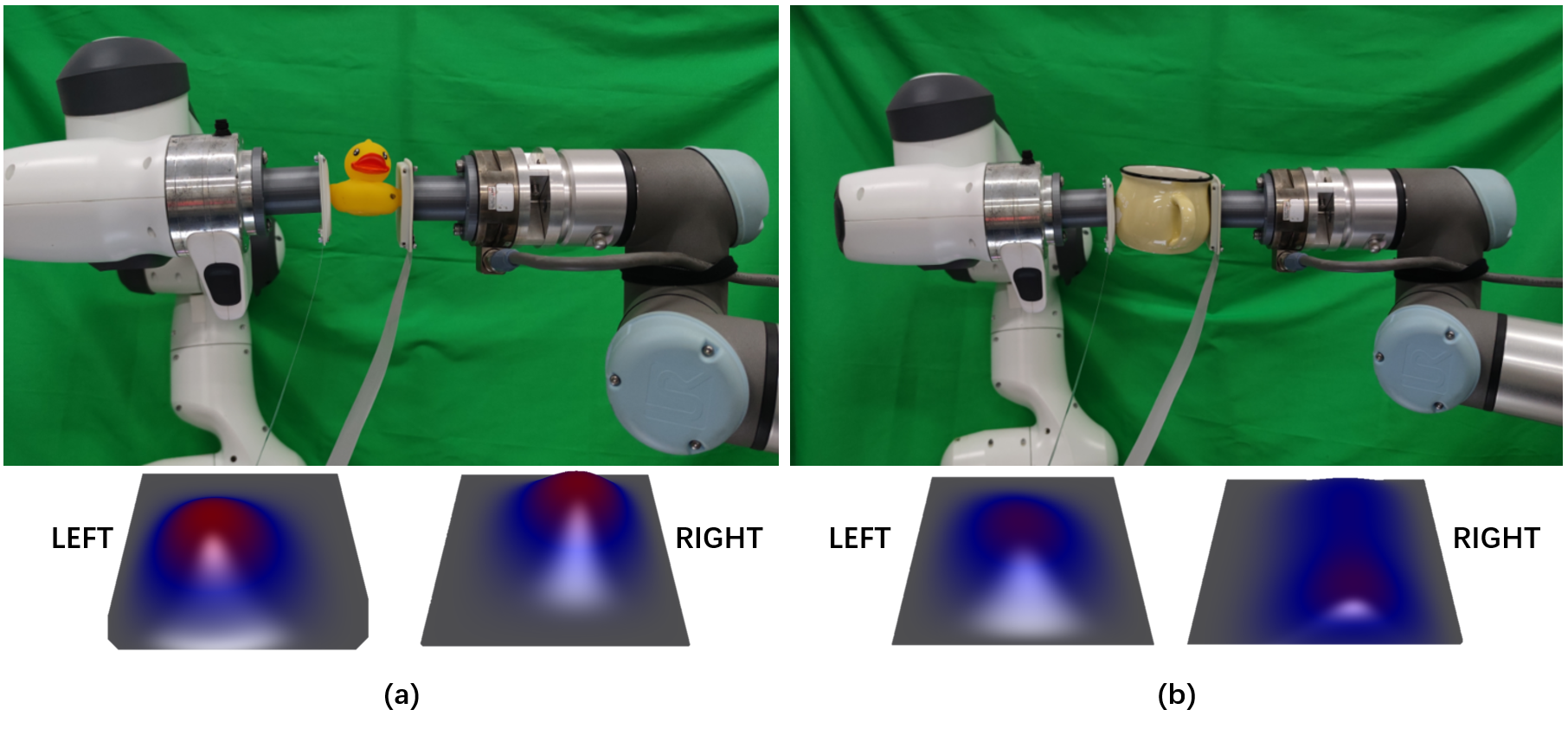}
    \caption{Bimanual tactile manipulation (a) Grasping a rubber object (b) Grasping a water cup. }.
    \label{arm}
    \vspace{-1cm}
\end{figure}

The experimental results demonstrate promising applications of the proposed tactile palm in robotic manipulation and object exploration. For example, as  discussed in \cite{billard2019trends}, one can use the palm as an intermediate step to change the state of the object, in order to accomplish the sequential more complex manipulation task, such as the in-hand manipulation task in \cite{andrychowicz2020learning}. These potential applications will be further investigated in our future work. In addition, from the details of the fabrication process, it can be known that more complex shaped tactile sensor can be designed. For example, a cylindrical shaped tactile pad can be developed to cover the link of a robot arm, which will possibly pave a new direction for whole body tactile sensing. One example would be bimanual tactile manipulation as shown in Fig.~\ref{arm}.


\section{Discussion and Conclusion}

In this paper, we proposed a novel biomimetic design of tactile palm. The tactile palm mainly consists of three parts: an array of electrodes, an elastomeric skin and the conductive sponge. The design and fabrication process are given in details, with a systematic method to estimate the contact position and the contact force. Both qualitative and quantitative experiments are conducted to demonstrate the effectiveness of the proposed design. In the future, we will further study the problem of how to combine the tactile palm with robot arm and fingers to accomplish more complex object manipulation tasks.


%
%
%
%

\bibliographystyle{style}
\bibliography{reference}
\end{document}